\documentclass[conference]{IEEEtran}

\IEEEoverridecommandlockouts
\usepackage{cite}
\usepackage{amsmath,amssymb,amsfonts}
\usepackage{algorithmic}
\usepackage{graphicx}
\usepackage{textcomp}
\usepackage{xcolor}
\usepackage{stfloats}
\usepackage{url}
\usepackage{verbatim}
\usepackage{threeparttable}
\usepackage{booktabs}
\usepackage{array, caption, threeparttable}
\usepackage[font=small,labelfont=bf,labelsep=none]{caption}
\usepackage[T1]{fontenc}
\usepackage{multirow}
\usepackage{pifont}
\usepackage{comment}
\usepackage{bm}

\def\BibTeX{{\rm B\kern-.05em{\sc i\kern-.025em b}\kern-.08em
		T\kern-.1667em\lower.7ex\hbox{E}\kern-.125emX}}
\begin{document}
	
	\title{
		
		A Self-supervised Contrastive Learning 
		
		Method for Grasp Outcomes Prediction
		\\
		
		\thanks{$^{1}$Guangdong Provincial Key Laboratory of Robotics and Intelligent System, Shenzhen Institute of Advanced Technology, Chinese Academy of Sciences, Shenzhen 518055, China. E-mail: \{cl.liu1, bh.huang, yw.liu, yz.su, k.mai, yp.zhang1, zk.yi, xy.wu\}@siat.ac.cn} 
		\thanks{$^{2}$University of Chinese Academy of Sciences, Beijing 100049, China.}
		\thanks{$^{3}$SIAT Branch, Shenzhen Institute of Artificial Intelligence and Robotics for Society, Shenzhen 518055}
		\thanks{$^{*}$Corresponding author: Zhengkun Yi, zk.yi@siat.ac.cn}
	}

	\author{\IEEEauthorblockN{Chengliang Liu$^{1,2}$, Binhua Huang$^1$, Yiwen Liu$^{1,2}$, Yuanzhe Su$^{1,2}$,\\ Ke Mai$^1$, Yupo Zhang$^1$, Zhengkun Yi$^{1,2,*}$, Xinyu Wu$^{1,2,3}$}}
	
	\maketitle
	
	\begin{abstract}
		In this paper, we probe the proficiency of contrastive learning techniques in forecast grasp outcomes, without supervision. Employing a dataset that's open to the public, we establish the effectiveness of contrastive learning techniques in accurately predicting grasp outcomes. More precisely, an impressive accuracy of 81.83\% is achieved by the dynamic-dictionary-based method combined with the momentum updating technique, using data from just one tactile sensor, surpassing other unsupervised techniques. Our findings underscore the promise held by contrastive learning techniques in the domain of robotic grasping, while emphasizing the critical role played by precise grasp prediction in securing stable grasps.
		
	\end{abstract}
	
	
	\section{Introduction}
	The field of robotic grasping has captivated researchers globally due to the escalating need for varied and skillful grasping and manipulation by industrial and service robots \cite{kwiatkowski2017grasp,fang2022tactonet}. The ability to predict grasp success based on sensory data collected prior to the grasp is a key factor in ensuring stable robotic grasping. Among the diverse sensory modalities available to robots, visual and tactile perceptions are often used because of their proven efficacy \cite{guo2017robotic}. Strategies based on computer vision and deep learning have shown significant advancements \cite{sundermeyer2021contact, mahler2019learning}, but they can sometimes fall short when robots operate in unstructured environments. Instances like suboptimal lighting conditions that could interfere with visual sensing \cite{yi2021touch}, obstructions in the actual working environment that obscure the target object \cite{phelan2017design}, or the necessity for sophisticated interaction with targets \cite{wang2021status}, can pose challenges.
	
	\begin{figure*}[htbp]
		\centerline{\includegraphics[scale=0.4]{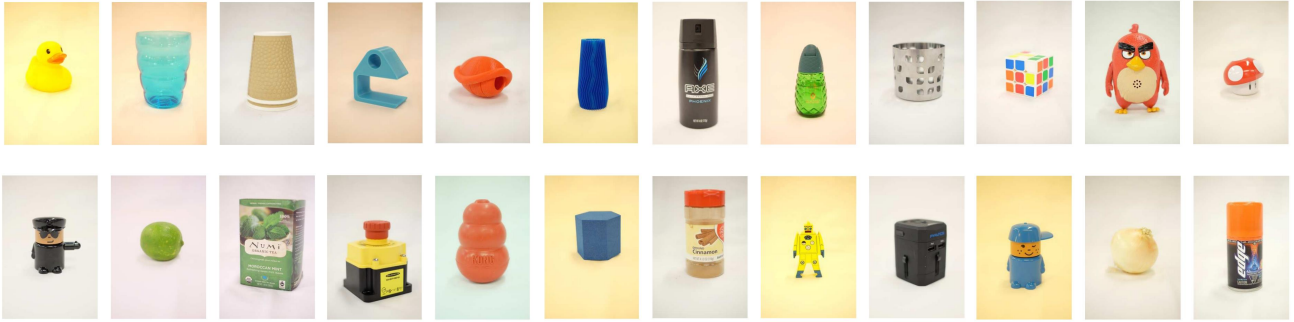}}
		\caption{. Examples of objects in Calandra dataset. Overall, 106 objects were included and 9296 grasps were performed in the dataset.}
		\label{物品}
	\end{figure*}
	
	Tactile perception stands out due to its direct access to interaction information. Lately, several significant studies have been based on tactile perception. For instance, Tulbure et al. \cite{tulbure2018superhuman} assembled a dataset of 36 common household materials with 100 samples each. They developed a TactNet-\uppercase\expandafter{\romannumeral2} network, an advancement on the original TactNet \cite{baishya2016robust}, and achieved a striking 95.0\% accuracy in material classification tasks using solely tactile data. Meanwhile, Zhou et al. \cite{zhou2022methods} introduced ordinal classification methods to discern different depths of hard inclusion concealed in silicone-based tissue phantoms, utilizing tactile data collected from a robotic palpation system. Yao et al. \cite{huang2021texture} gathered tactile data by sliding a biomimetic tactile sensor over ten different materials and designed a convolutional neural network capable of reaching 98.5\% test accuracy. In their work, Mi et al. \cite{mi2021tactile} proposed two graph convolution network (GCN) models for predicting grasp outcomes. They converted the electrode values of BioTac sensors into graph representations, achieving commendable performance. Cockbum et al. \cite{cockbum2017grasp} delved into identifying common factors shared by successful and failed grasp attempts, which could help robots predict the outcome of any object's grasp. Their method uses sparse coding and straightforward support vector machines to distill the most pertinent high-level features from the source data for predicting grasp outcomes. However, the former approach's data processing method into graph representations is severely constrained by the source data format, and the latter method is viable only when the tactile data is of low dimensionality.
	
	Calandra et al. \cite{calandra2017feeling} created a visual and tactile model and performed experiments showing that touch feedback greatly improves grasp prediction. Their model fused data from vision and touch by joining high-dimensional features. Cui et al. \cite{cui2020self} proposed a new method to combine visual and tactile data using self-attention. They formed fusion features by combining 2D deep features from each modality and applying self-attention. In this way, Cui et al. obtained a better accuracy and they used a training set of similar size to the former work. Yang et al.\cite{yang2018predict} utilized LSTM to extract tactile features and combined these with visual and weight features extracted via deep neural networks. Their fused model achieved promising results.

	While the techniques suggested in these studies can achieve respectable results in predicting grasp outcomes, they don't explore how to maintain high performance using self-supervised learning methods. Self-supervised learning holds substantial importance in artificial intelligence because of its capacity to draw useful insights from vast data sets. This is particularly useful for numerous applications, as real-world scenarios often provide us with extensive data, but we might not always have access to human-annotated training data. One efficient method of self-supervised learning is contrastive learning. In unsupervised comparative learning methods, self-supervised learning is usually achieved by constructing pretext tasks to generate pseudo labels for unlabeled data. Gidaris et al.\cite{gidaris2018unsupervised} first defined a small set of discrete geometric transformations. After that, the feature pairs obtained by projecting two images from the same image into the feature space after different geometric transformations were defined as positive pairs, and the feature pairs obtained by projecting images from different images into the feature space after geometric transformations were defined as negative pairs. They constructed a model that mapped positive pairs closer together and negative pairs further apart and closed the gap with supervised feature learning in this simple way. Oord et al. \cite{oord2018representation} introduced a framework for obtaining condensed latent representations, which integrated autoregressive modeling and noise-contrastive estimation with insights from predictive coding to learn abstract representations without supervision. Wu et al. \cite{wu2018unsupervised} used the memory bank mechanisms to build dynamic dictionaries and the contrastive loss was applied to the tokens sampled from the dictionaries. He et al. \cite{he2020momentum} enhanced the approach by constructing a dynamic dictionary utilizing a queue and a moving-average encoder.
	
	In previous works, positive pairs in contrastive learning methods have been composed of samples that are similar at the level of contours, such as the signature of the same person, the face of the same person, or different views of the same item. However, in this article, we propose a novel approach by considering pre-grasp tactile data with a successful simultaneous grasp or a failed simultaneous grasp as a pair of anchor and positive samples.
	
	Our work stands out because the tactile grasp samples collected using Gelsight tactile sensors, although also image-based, provide only very localized contour details about the target object. Furthermore, grasp samples for the same label can exhibit substantial variability, as the success or failure of a grasp trial is not easily depicted using pre-grasp tactile surface information.
	
	Our contributions are summarized as follows:
	\begin{itemize}
		\item [1)] 
		We investigate the effectiveness of self-supervised contrastive learning methods on a Gelsight-based tactile dataset and verify that the contrastive learning methods perform well with Gelsight-based tactile datasets.
		\item [2)]
		We successfully migrated the implementation of the MoCo algorithm on the grasp outcomes prediction task which achieves $81.83\%$ predictive accuracy with a single tactile sensor data.
		\item [3)]	
		We validate that the self-supervised contrastive learning algorithm MoCo outperforms other contrastive learning methods and the autoencoder method through experiments on the Calandra public dataset.
	\end{itemize}

	\section{Methodology}
	
	In this section, we initially provide a detailed introduction to the Calandra Grasping dataset in Section II-A. Subsequently, we comprehensively describe the proposed architecture in Section II-B.
	
	\subsection{Calandra Grasping Dataset}
	
	Calandra et al. \cite{calandra2017feeling} compiled a multimodal dataset that includes both visual and tactile data, featuring 9296 grasp samples from 106 distinct objects. The Gelsight sensor was used to capture tactile data, while the Microsoft Kinect sensor was used to collect visual data. Specifically, two Gelsight sensors were affixed to the fingers of a Weiss WSG-50 robotic arm, and a stationary Microsoft Kinect camera was positioned in front of the grasping platform to take pictures. During each grasp, they collected visual and tactile data at three specific points in time: $T_a$, when the robotic arm is in the initial position before the grasp; $T_b$, when the fingers of the robotic arm have grasped the object and remained stable; and after the lift $T_c$, a few seconds after the robotic arm has completed the lift-off procedure.
	
	In this paper, as shown in Figure 1, the Gelsight sensor on the left is referred to as Gelsight-L, and the one on the right is referred to as Gelsight-R. The tactile data captured by Gelsight-L at time $T_a$ is denoted as $G_{La}$, and similarly for other time points. It's important to emphasize that the dataset only contains RGB data captured by the Gelsight and Kinect sensors, and in this paper, only the data from times $T_a$ and $T_b$ are used predicting grasp outcomes.
	
	\subsection{The Overall Framework}
	
	The proposed overall framework is basically constructed based on MoCo v2 \cite{chen2020improved}, which is an improved version of MoCo v1\cite{he2020momentum} with the addition of some of the tricks used in SimCLR\cite{chen2020simple}. The process of contrastive learning involves the identification of anchor, positive, and negative samples through predefined criteria. The anchor and positive samples are considered as similar pairs, while the anchor and negative samples are considered as dissimilar pairs. The objective is to optimize the model such that the representations of similar pairs are embedded closely, while those of dissimilar pairs are embedded far apart in the feature space\cite{oord2018representation}. In this paper, we follow a simple instance discrimination task\cite{Wu_2018_CVPR,Ye_2019_CVPR}, where images obtained from a single image through various data augmentation techniques are considered as similar pairs, and images from different instances are considered as dissimilar pairs. Our framework comprises the following three major components.
	
	\begin{figure*}[htbp]
		\centerline{\includegraphics[scale=0.78]{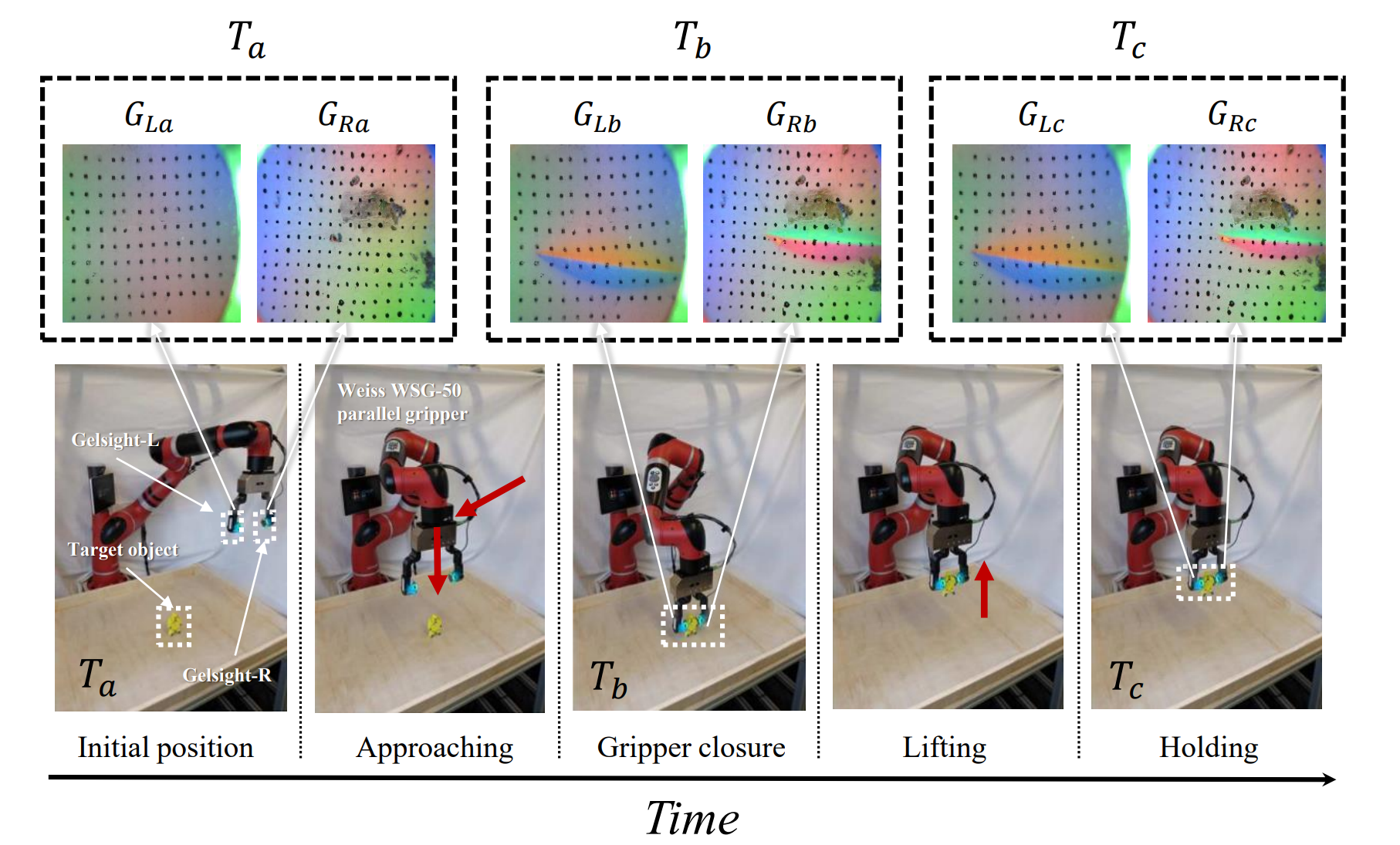}}
		\caption{. Timeline of a data collection trial illustrating various phases of grasping. Two Gelsight sensors were affixed to the fingers of a Weiss WSG-50 robotic arm, and a stationary Microsoft Kinect camera was positioned in front of the grasping platform to take pictures. During each grasp, they collected visual and tactile data at three specific points in time: $T_a$, when the robotic arm is in the initial position before the grasp; $T_b$, when the fingers of the robotic arm have grasped the object and remained stable; and after the lift $T_c$, a few seconds after the robotic arm has completed the lift-off procedure. The Gelsight sensor on the left is referred to as Gelsight-L, and the one on the right is referred to as Gelsight-R. The tactile data captured by Gelsight-L at time $T_a$ is denoted as $G_{La}$, and similarly for other time points. This figure is adapted from [28].}
			
		\label{Chronology_of_a_data}
	\end{figure*}
	
	\begin{figure}[htbp]
		\hspace{-0.2cm}\centerline{\includegraphics[scale=0.43]{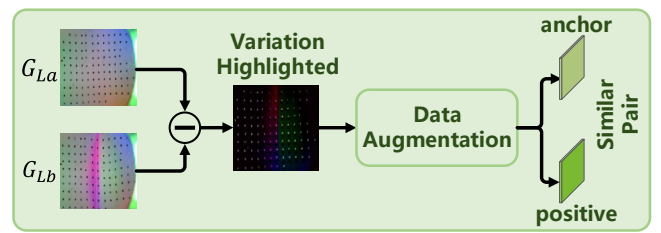}}
		\caption{. Data preprocessing. Firstly, we subtract $G_{La}$ from $G_{Lb}$ to highlight the area of variation due to the contact with the target object \cite{calandra2017feeling}. Then $G_{La}-G_{Lb}$ undergoes multiple data augmentation to generate two images, and them are viewed as a similar pair. One of them is considered as the anchor sample, and the other one is considered as the positive sample.} 
		\label{similar_pair}
	\end{figure}
	
	\begin{figure*}[htbp]
		\hspace{-0.2cm}\centerline{\includegraphics[scale=0.78]{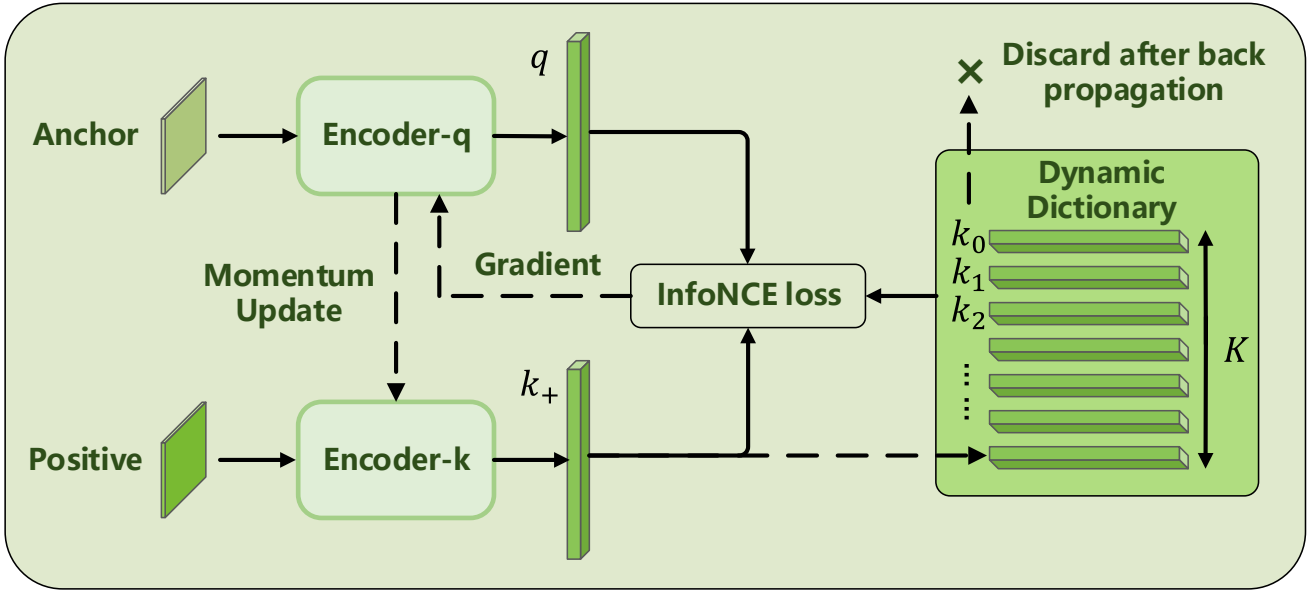}}
		\caption{. Illustration of dynamic dictionary and momentum update. The feature extracted from the anchor sample is denoted as $q$, and that extracted from the positive sample is denoted as $k_+$. The features stored in the dynamic dictionary are denoted as $k_i$ for i in the range of 0 to K-1, where K denotes the capacity of the dictionary. The InfoNCE loss is computed using $q$, $k_+$, and $k_i$ (for i in the range of 0 to K-1) and the parameters in the encoder-k are detached during the process. The encoder-q is updated end-to-end by back-propagation and the encoder-k is updated by momentum update. Then, the oldest key in the dynamic dictionary is dequeued, and the $k_+$ is enqueued into the dictionary.} 
		\label{dynamic_dict}
	\end{figure*}
	
	$\bullet$ Data Preprocessing and Augmentation. Realizing the presence of abundant redundant information in $G_{Lb}$, we subtract $G_{La}$ from $G_{Lb}$ to emphasize the area that varies due to contact with the target object. For data augmentation, we first resize the images to be $256\times256$ and randomly sample $224\times224$ crops from them. Next, random color jittering, random horizontal flip, random grayscale conversion, and random gaussian blur are applied. As illustrated in Fig. \ref{similar_pair}, we generate two different versions of images from one image by the data augmentation techniques mentioned above and consider them as a similar pair. One of them is viewed as an anchor sample and the other one is viewed as a positive sample.
	
	$\bullet$ Encoder. The encoder module within our framework is composed of two elements: the backbone and the MLP (multi-layer perceptrons) nonlinear projection head. The input image is initially processed through convolutional neural backbone networks, like Resnet\cite{he2016deep}, VGG\cite{simonyan2014very}, and so on, to extract their latent features. According to a study by Chen et al. \cite{chen2020simple}, it has been found that the use of a learnable nonlinear projection head can significantly enhance the quality of learned representations. In our study, we implemented this concept using a two-hidden-layer MLP.
	
	$\bullet$ Momentum Contrast. In contrastive learning, it is important to construct and contrast similar pairs and dissimilar pairs as many as possible. From this point of view, a smart solution is to maintain a large dictionary which contains abundant features from all samples \cite{he2020momentum,wu2018unsupervised}. In this paper, we construct a dynamic dictionary following \cite{he2020momentum}.
	
	Specifically speaking, the dynamic dictionary is maintained as a queue of features of data samples. The features of a mini-batch will be enqueued as the keys in the dictionary. Meanwhile the oldest features will be discarded during training \cite{he2020momentum}. By utilizing this method, the encoded features can be reused in subsequent contrastive learning while minimizing the cost of storing and maintaining the dictionary as the stored features are encoded and of low dimensionality.
	
	We denote the feature extracted from the anchor sample as $q$, and that extracted from the positive sample as $k_+$, and the features stored in the dynamic dictionary as $k_i$ \{$i=0,1,2,...,K-1$\}, where $K$ denotes the capacity of the dictionary. With similarity measured by dot product, a form of a contrastive loss function, known as InfoNCE \cite{oord2018representation} is employed:
	\begin{equation}
		\label{Equa.1}
		\tag{1}
		\mathcal{L}_q=-\log\frac {\exp(q\cdot k_+/\tau)} {\sum_{i=0}^{K}\exp (q\cdot k_i/\tau)},
	\end{equation}
	where $\tau$ is a temperature hyper-parameter \cite{wu2018unsupervised}. In the above way, the data samples of one minibatch are compared with the features stored in the whole dictionary every time the gradient back-propagation is performed by calculating the loss $\mathcal{L}_q$. 
	
	As illustrated in Fig. \ref{dynamic_dict}, it needs to be emphasized that $q$ and $k_i$ are obtained using different encoders, i.e. encoder-$q$ and encoder-$k$. To ensure that the features in the dictionary remain consistent and updating, encoder-$k$ is updated with momentum instead of simply copying the parameters from encoder-$q$ \cite{he2020momentum}. We denote the parameters of encoder-$q$ as $\theta_q$ and those of encoder-$k$ as $\theta_k$, and $\theta_k$ is updated formally as below:
	
	\begin{equation}
		\label{Equa.2}
		\tag{2}
		\theta_k\leftarrow m\theta_k+(1-m)\theta_q,
	\end{equation}
	
	where $m\in [0,1)$ is a momentum coefficient. By setting the momentum hyperparameter $m$, $\theta_k$ can remain more stable with updating. 
	
	\section{Experiment}
	
	In this section, to verify the proposed architecture, we first implement experiments on the Calandra dataset and study the performance compared with triplet net \cite{hoffer2015deep}, memory bank \cite{wu2018unsupervised}, and autoencoder\cite{hinton2006reducing}. Furthermore, we analyze the experimental results. 
	
	\subsection{Implementation Details}
	The learning rate was set to $1\times10^{-2}$ initially and was decayed to $1\times10^{-6}$ in 200 epochs with the cosine decay schedule without restarts\cite{loshchilov2016sgdr}. We train at batch size 200 on a single GeForce GTX 3090 GPU with PyTorch. The momentum updating parameter $m$ was set to $0.999$. The temperature hyper-parameter was set to $0.07$ \cite{he2020momentum}, and the number of dimensions of latent features was set to $128$. 
	
	In the subsequent experiments, we utilize 6,000 non-overlapping samples from the Calandra dataset \cite{calandra2017feeling} as the training set and 3,000 non-overlapping samples as the test set. Although there may be duplicate items present in both sets, this partitioning ensures a fair assessment of the model's performance in predicting grasp outcomes on unseen data. The capacity of the dynamic dictionary $K$ was set to 5800 to prevent keys encoded from the image from being the same as the current anchor samples. It should be noted that, according to He et al. \cite{he2020momentum}, the $K$ can actually be set to a much bigger number than the number of training set samples considering the percentage of incorrect keys in the dictionary is so small that it can be ignored.
	
	\subsection{Evaluation of the Algorithm Performance}
	
	To evaluate the effectiveness of the proposed method, we compare it with the contrastive learning methods using a memory bank \cite{wu2018unsupervised}, the triplet net \cite{hoffer2015deep}, and the autoencoder method\cite{hinton2006reducing}. In the following experiments, these methods are used with Resnet50 as the backbone, and they are briefly described below.
	
	\begin{itemize}
		\item[$\bullet$] Memory Bank \cite{wu2018unsupervised}. The primary distinction between a memory bank and the dynamic dictionary utilized in MoCo\cite{he2020momentum} is their capacity and method of updating. Specifically, the capacity of a memory bank is equivalent to the number of samples the training dataset contains, and the feature keys and samples are in a one-to-one correspondence. In contrast, the capacity of the dynamic dictionary in MoCo\cite{he2020momentum} is not restricted in this manner. Additionally, the keys within a memory bank are updated in accordance with the most recent encoder parameters, and can be considered a specific instance where the momentum updating parameter $m$ is set to 0 in MoCo.
		\item[$\bullet$] Triplet net \cite{hoffer2015deep}. The triplet net is one of the basic methods of contrastive learning. It constructs triplet samples, which include two samples with the same labels and one sample with a different label from them. We use it and employ the triplet margin loss \cite{schroff2015facenet}.
		\item[$\bullet$] Autoencoder \cite{hinton2006reducing}. An autoencoder refers to a particular neural network design primarily employed for unsupervised learning. It is composed of two key parts: an encoder and a decoder. The encoder translates the input data into a compressed format, often referred to as the bottleneck or latent representation, and the decoder then transforms this compact format back into the original input space. The autoencoder's objective is to learn a succinct yet informative representation of the input data by reducing the reconstruction error between the original input and the decoder's output. Such a mechanism proves beneficial for tasks like reducing dimensionality, compressing data, and detecting anomalies.
		
	\end{itemize}
	
	To assess the quality of the representations generated by these methods, we employ a suite of evaluation protocols, specifically utilizing three classifiers: K-Nearest Neighbors (KNN), Support Vector Machine (SVM), and Multilayer Perceptron (MLP) to predict grasp outcomes using the representations. As per Lall and Sharma \cite{lall1996nearest}, $k$ ought to be $\sqrt{n}$ for $n>100$, where $n$ represents the number of training samples. Consequently, we set $k=\sqrt{6000}\approx77$ in our experiments. The evaluation process involves determining the accuracy of the models in predicting the success or failure of grasps using the data from the left single-finger tactile perception. The experimental results are presented in Table \ref{Table1}, with the highest accuracy highlighted in bold, and it is observed that the MoCo model achieves higher prediction accuracy than the other unsupervised learning models. 
	
	\begin{table}[t!]
		\begin{center}
			\tabcolsep=0.35cm
			\renewcommand\arraystretch{1.2}
			\caption{. Classification accuracy rate (\%) of the models when using the \textbf{Left} tactile Gelsight sensor data on Calandra dataset. The highest accuracies of unsupervised learning models in the same classifier are highlighted in bold.}
			\captionsetup[table]{
				labelsep=newline,
				singlelinecheck=false,
			}
			\begin{tabular}{cccc} 
				\cline{1-4} 
				\textbf{Model} & \textbf{KNN} & \textbf{SVM} & \textbf{MLP}\\
				
				\cline{1-4} 
				
				Autoencoder & 74.23\% & 73.90\% & 75.20\% \\
				Triplet net & 75.33\% & 75.10\% & 75.88\%\\
				Memory bank & 78.30\% & 78.43\% & 78.67\% \\
				MoCo & \pmb{80.37\%} & \pmb{80.83\%} & \pmb{81.83\%} \\
				
				\cline{1-4} 
				
			\end{tabular}
			\label{Table1}
		\end{center}
	\end{table}

	\section{Discussion and Conclusion}
	
	In this paper, we investigate the performance of contrastive learning methods for predicting grasp outcomes in an unsupervised manner using a Gelsight-based tactile dataset. Our algorithm, which is based on MoCo, is designed for predicting grasp outcomes using pre-grasp information from a single Gelsight tactile sensor. It achieves a prediction accuracy of 81.83\% on the Calandra dataset, surpassing the performance of other unsupervised learning algorithms.
	
	Our results demonstrate the potential of contrastive learning methods for applications in the field of robot grasping. However, there are some potential limitations of the proposed method. One drawback of the MoCo algorithm is its instability during training, which can lead to a slight decrease in performance rather than a failure to converge. To address this issue, freezing the patch embedding layer's parameters or employing gradient clipping can improve training stability. Another drawback is the requirement for a large batch size to observe enough negative samples, which can lead to dip problems when using a larger batch size. 
	
	In future work, we intend to further explore the potential of contrastive learning methods for targeted optimization in this domain in order to improve the performance of robot grasping systems.
	
	\section*{Acknowledgement}
	
	The work described in this paper is partially supported by the National Natural Science Foundation of China (No. U22A2056 and No. 62003329), partially supported by the Science and Technology Innovation Commission of Shenzhen (No. JSGGZD20220822095401004), and partially supported by the National Natural Science Foundation of Guangdong Province of China (No. 2021A1515011316)
	
	\bibliographystyle{IEEEtran}
	\bibliography{references}

\end{document}